# Evaluating Fine-Tuned and Base Language Models in Maternal and Vaccination Healthcare for African Settings

**AUTHORS (Surname, Firstname) :**

- AJIBADE, ABDULQUDDUS (1,2) - Contact person (tech@helpmum.org)
- ODUNSI, OLUWASEUN (1)
- ANIMASAUN, IYINOLUWA (1)
- NWAKANMA-AKANNO, CHIOMA (1)
- OGUNTUASE, OLUWASEGUN (1)
- AKINBUWA, OLUWAFUNKE (1)
- ADERENI, ABIODUN (1) - Contact person (biodun@helpmum.org)

AFFILIATIONS

**(1)** HelpMum, Nigeria

# Evaluating Fine-Tuned and Base Language Models in Maternal and Vaccination Healthcare for African Settings

## I. Abstract

**Background:** Large language models (LLMs) can improve healthcare information delivery in low-resource settings but may produce inaccurate or culturally inappropriate advice. This study evaluated domain-specific fine-tuning for maternal health and vaccination in Nigeria.

**Objective:** To compare HelpMum's MamaBot-Llama and Vax-Llama with Meta's Llama-3.1-8B-Instruct for accuracy, safety, clarity, contextual appropriateness, and trustworthiness.

**Methods:** We evaluated 200 healthcare questions, 100 each for maternal health and vaccination, across five subdomains per domain. MamaBot-Llama and Vax-Llama were fine-tuned using Low-Rank Adaptation on over 36,000 maternal health and 9,000 vaccination question-answer pairs, respectively. Two Nigerian licensed physicians independently rated responses using a 5-point Likert scale. Paired comparisons used Wilcoxon signed-rank tests.

**Results:** Performance varied by domain. MamaBot-Llama significantly outperformed the base model across all criteria, with a 4.9% overall improvement ($p < .001$), including gains in clinical trustworthiness (+7%) and medical accuracy (+5%). Critical issues decreased by 50%, and clinicians preferred it in 78% of cases. In contrast, Vax-Llama showed a 5.2% overall decline ($p < .001$), with critical issues increasing by 192% and safety concerns by 400%.

**Conclusions:** Domain-specific fine-tuning can improve healthcare LLM performance when based on high-quality, clinician-curated data, but may also degrade performance when dataset quality is inadequate. Rigorous domain-specific validation is essential before clinical deployment. Physician evaluators provided informed consent, and chatbot logs were anonymized.

**Keywords:** Large language models; Fine-tuning; Maternal health; Vaccination; Healthcare AI; Low-resource settings; Nigeria; Model evaluation; LoRA; Medical accuracy

# Evaluating Fine-Tuned and Base Language Models in Maternal and Vaccination Healthcare for African Settings

## II. INTRODUCTION

### A. Background

Large language models (LLMs) have rapidly become influential in healthcare information delivery. The public release of AI chatbots like ChatGPT in 2022 led many people to seek medical advice from these systems. By 2024, roughly 17% of adults reported using AI chatbots at least once a month for health information (Presiado et al., 2024). These models leverage vast medical knowledge and can perform impressively on clinical tasks. For instance, Google's Med-PaLM 2, a large model fine-tuned for medicine, achieved expert-level performance (about 86.5% accuracy) on U.S. medical exam questions (Singhal et al., 2023). Such successes illustrate the impact of LLMs in healthcare. However, generic models also carry risks. They can hallucinate false but plausible answers (Alkaissi & McFarlane, 2023), and their responses may be inappropriate for certain cultures or contexts. A model trained mostly on Western data might give advice that is irrelevant or insensitive in an African setting. These limitations highlight the need for LLMs that are carefully adapted to the medical domain and local context to ensure information is accurate, safe, and culturally relevant.

### B. Importance of Maternal Health & Vaccination

Maternal health is a critical area where accurate, context-aware information can be life-saving. This is especially true in low-resource settings with high maternal risks. Nigeria, for example, has one of the world's worst maternal mortality rates, accounting for about 28.5% of global maternal deaths (EQUAL Research Consortium, 2023). Many of these deaths are preventable with timely care and correct knowledge. Inadequate health literacy contributes to the problem; one study found over 40% of Nigerian mothers had low understanding of maternal health, which was associated with poorer use of antenatal services (Bello et al., 2022). This underscores the importance of equipping expectant mothers with trusted guidance tailored to their needs and environment.

Vaccination is another vital aspect of maternal and child health where information is crucial. Immunization programs can falter when misinformation spreads. In Nigeria, false rumors about vaccines (for example, the myth that vaccination causes infertility) have made many families hesitant. A recent study confirmed that such misinformation significantly lowers vaccine uptake among pregnant women (Adeyanju et al., 2022). Delivering relevant and reliable advice on pregnancy care and immunizations can empower women in underserved communities, helping them make informed decisions, embrace healthy practices, and ultimately reduce the risks for both mother and child.

## C. Challenges in AI-Generated Medical Advice

Using AI to provide medical information poses several challenges that must be addressed to ensure quality and safety. The first is accuracy. An AI model's answer must align with established medical knowledge; any incorrect recommendation can lead to harm or erode trust. Yet LLMs do not guarantee correctness; they sometimes output confidently worded misinformation (Alkaissi & McFarlane, 2023). If not caught, such errors could mislead patients or health workers (e.g., suggesting an unsafe home remedy or a wrong medication dose). This risk of "errors with confidence" means rigorous validation and oversight are needed when applying LLMs in healthcare. Relatedly, maintaining user trust is critical: if people suspect an AI's answers are unreliable, they will not use it. At present, public confidence in AI health advice is limited; only about 29% of people in one survey said they would trust information from a chatbot for health matters (Presiado et al., 2024). Ensuring consistent accuracy and clearly communicating the AI's limitations are key steps to improve trustworthiness.

Another challenge is cultural relevance. A one-size-fits-all model may give advice that is not feasible or acceptable in a particular community. For instance, an LLM might recommend a diagnostic test or diet that is unavailable or unfamiliar in a rural Nigerian setting. It might also overlook local beliefs and customs that affect healthcare decisions. Such mismatches can render the advice ineffective. In practice, this requires fine-tuning models with region-specific data (including local languages, common practices, and epidemiological patterns) so that their responses resonate with users. Addressing both the accuracy issue and the cultural/contextual alignment issue is essential before AI-generated advice can be safely deployed for health education in places like Nigeria.

### D. Study Rationale & Objectives

The deployment of large language models (LLMs) in low-resource healthcare settings, such as Nigeria, remains underexplored, with most research focusing on general models in high-income contexts. This study addresses this gap by evaluating the performance of fine-tuned models , tailored to maternal healthcare and vaccination domains, against the base Meta's Llama-3.1-8B-Instruct model, in delivering high-quality responses. The domain-specific models were fine-tuned using datasets derived from trusted sources, including WHO, UNICEF, and HelpMum's MamaBot and Vax AI chatbot conversations, ensuring relevance to African healthcare contexts. The primary objective is to compare the performance of these fine-tuned models with the base model in providing medically accurate, safe, contextually appropriate, and clear responses, as evaluated by medical experts.This study aims to demonstrate the value of culturally adapted, domain-specific LLMs in supporting maternal healthcare and vaccination guidance in low-resource settings. Successful outcomes would underscore the importance of fine-tuning LLMs with localized medical knowledge to improve the reliability and relevance of AI-driven health information delivery in underserved communities.

## III. Methodology

### A. Study Design:

This study utilized a cross-sectional, comparative experimental design to evaluate the performance of two domain-specific fine-tuned large language models (LLMs) against a base

# Evaluating Fine-Tuned and Base Language Models in Maternal and Vaccination Healthcare for African Settings

model in addressing healthcare queries relevant to maternal health and vaccination in African contexts. The evaluation process involved generating responses to domain-specific questions, conducting blinded expert evaluations by medical professionals, and analyzing results using appropriate statistical methods. The study aimed to determine whether fine-tuning enhances the models' ability to provide medically accurate, safe, contextually appropriate, clear, and trustworthy responses tailored to African healthcare settings, with particular focus on Nigerian contexts.

## B. Participants

Two Nigerian licensed Nigerian physicians served as independent evaluators for all model responses. Both evaluators provided verbal informed consent prior to starting the evaluation. The study purpose, task nature, publication of aggregated ratings, and the right to withdraw were explained by the principal investigator. Both evaluators are co-authors of this manuscript; their dual role was disclosed and accepted. Their individual identities remained known to the research team, but all ratings were anonymized during analysis and presentation.

## C. Materials

### 1. Base Model Configuration

The base model compared was Meta-Llama-3.1-8B-Instruct (Meta AI, 2024), an 8-billion parameter instruction-tuned large language model (LLM) optimized for general-purpose reasoning, comprehension, and dialogue generation across a wide range of tasks. It was pretrained on a large corpus of multilingual and multimodal data, enabling strong contextual understanding and coherent text generation. As an instruction-following variant, it is specifically aligned for natural interaction, factual consistency, and safety in response generation.

### 2. Fine-Tuned Models

# Evaluating Fine-Tuned and Base Language Models in Maternal and Vaccination Healthcare for African Settings

Two specialized models were developed through domain-specific fine-tuning:

I. **MamaBot-Llama Model**: Fine-tuned on the maternal health dataset for pregnancy-related queries
II. **Vax-Llama Model**: Fine-tuned on the vaccination dataset for immunization-related queries

## D. Data Collection and Preprocessing

### 1. Maternal Health Dataset

The maternal health dataset comprised over 36,000 question-answer pairs aggregated from three primary sources: LLM-augmented web-scraped contents, clinician-curated data and conversation logs from MamaBot AI.

Web-scraped content was collected from authoritative sources including the World Health Organization (WHO) maternal health pages, UNICEF child health resources, and African health portals such as Nigeria Health Watch. Web scraping was performed using an asynchronous script with *aiohttp* and *BeautifulSoup* libraries for HTML parsing, using *Trafilatura* with *favor_precision=True* for high-quality text extraction. Internal links were followed recursively to expand the corpus. GPT-4o was then used to generate question-answer pairs from the raw text scraped from the websites using few-shot prompting strategies. Prompts were specifically tailored to Nigerian mothers (e.g., "Generate 5 distinct questions and answers about maternal healthcare from a Nigerian perspective"). All scraping respected the robots.txt directives of each website, implemented a 1‑second delay between requests, and accessed only publicly available pages. No login credentials or anti‑scraping circumvention were used.

Clinician-curated pregnancy guides were compiled by a Nigerian licensed medical doctor, providing detailed week-by-week pregnancy information covering weeks 1-40, including fetal development, maternal health, nutrition, and exercise recommendations. Approximately 32% of the dataset consisted of clinician-curated content, enhancing overall reliability.

# Evaluating Fine-Tuned and Base Language Models in Maternal and Vaccination Healthcare for African Settings

Additionally, anonymized conversation logs from the HelpMum's MamaBot AI chatbot (a Whatsapp-based AI chatbot for maternal and child healthcare) containing real-world user queries and responses on maternal health were formatted into question-answer pairs. Data preprocessing involved deduplication and normalization procedures.

**2. Vaccination Dataset**

The vaccination dataset included over 9,000 question-answer pairs sourced from 2 sources: LLM-augmented web-scraped contents and conversation logs from Vax-AI.

Similar to the maternal health data, web-scraped content was collected from authoritative sources including WHO and UNICEF immunization pages, CDC Nigeria vaccination guidelines from the Nigeria Primary Health Care Development Agency, and other vaccination-related authoritative websites. The same web scraping methodology employed for the maternal health data was also used here. The raw web data were processed into question-answer pairs using GPT-4o with prompts emphasizing appropriate terminologies (e.g., "ORAL POLIO VACCINE" instead of "OPV") and resource-constrained settings. Approximately 89% of the dataset was created using this LLM-augmented method.

Anonymized conversation logs from the HelpMum's Vax-AI chatbot (a Nigerian-specific Whatsapp-based AI chatbot for vaccination and immunization queries) provided real-world vaccination queries formatted into question-answer pairs.

# Evaluating Fine-Tuned and Base Language Models in Maternal and Vaccination Healthcare for African Settings

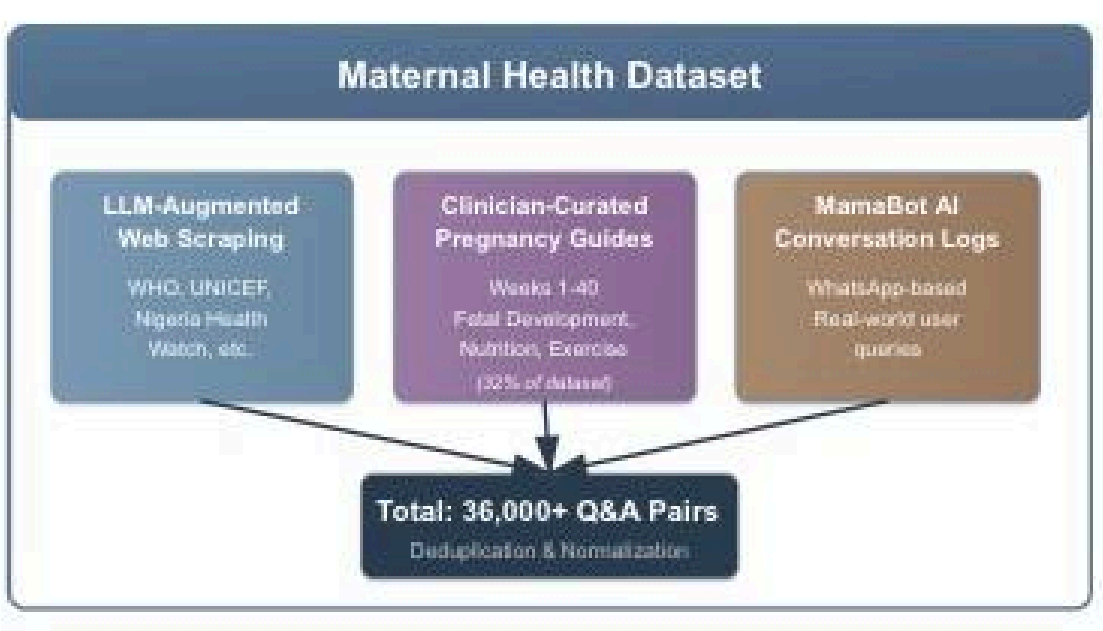

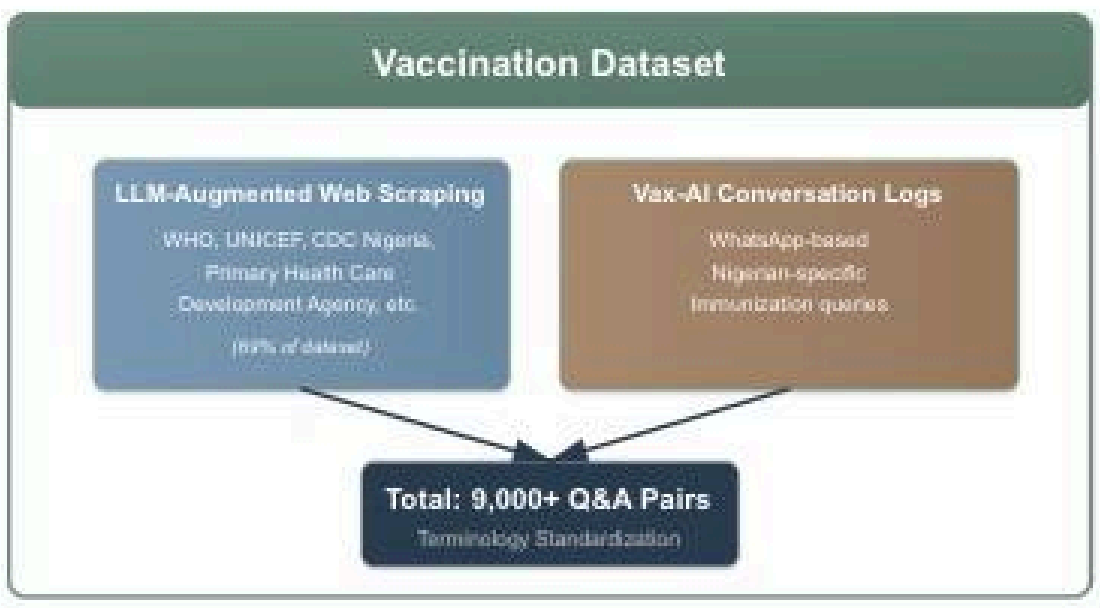

**Figure 1:** *Data collection and preprocessing pipeline. Web scraping, clinician-curated guides, and chatbot logs were processed into structured question-answer pairs.*

### 3. Dataset Preparation

Both datasets were divided into training (90%) and evaluation (10%) subsets. Text was formatted into chat templates following the structure: *[{"role": "user", "content": "..."}, {"role": "assistant", "content": "..."}]* to ensure compatibility with model inputs. Text segments were chunked to a maximum of 3,000 tokens using *tiktoken* before processing to ensure manageable input sizes.

## E. Model Training Procedure

### 1. Fine-Tuning Configuration

# Evaluating Fine-Tuned and Base Language Models in Maternal and Vaccination Healthcare for African Settings

Both models employed the Low-Rank Adaptation (LoRA) methodology for efficient fine-tuning of large language models. LoRA introduces low-rank decomposition matrices into specific attention and feed-forward layers, allowing the model to learn task-specific adaptations without updating the full set of base parameters. This significantly reduces computational cost and memory requirements while maintaining high performance.

The following configuration was used for both fine-tuned models: *r = 16, lora_alpha = 32, target_modules = ['q_proj', 'v_proj', 'up_proj', 'down_proj', 'gate_proj', 'k_proj', 'o_proj'], lora_dropout = 0.05, and task_type = "CAUSAL_LM"*. These parameters were selected to balance representational efficiency and stability during training. The *target_modules* were chosen to capture both attention and projection layers, ensuring adaptation across key components of the transformer architecture. This approach enabled parameter-efficient fine-tuning while preserving the base model's generalization capability, making it well-suited for domain-specific adaptation in resource-constrained settings.

Training arguments were configured as follows: *per_device_train_batch_size=1, per_device_eval_batch_size=1, gradient_accumulation_steps=2, num_train_epochs=3, learning_rate=2e-4, optim="paged_adamw_32bit", max_seq_length=512, evaluation_strategy="steps", eval_steps=0.2, logging_steps=1, warmup_steps=10,* with progress reporting to Weights & Biases (*wandb*) and model deployment to Hugging Face Hub.

### 2. Training Execution

The base model was loaded using *AutoModelForCausalLM.from_pretrained* with specified quantization settings and *device_map='auto'*. The corresponding tokenizer was loaded using *AutoTokenizer.from_pretrained.* The *SFTTrainer* from the Hugging Face TRL library facilitated supervised fine-tuning. Post-training, LoRA adapters were merged with the base model using *model.merge_and_unload()*.

Metrics for the Vax-Llama model included an evaluation loss of 0.7393, an evaluation runtime of 191.51 seconds, a training loss of 1.8186, and a training gradient norm of 4.3407. For the MamaBot-Llama model, the training and validation losses were 0.4654 and 0.5168, respectively. These results indicate consistent training stability and convergence across both models.

# Evaluating Fine-Tuned and Base Language Models in Maternal and Vaccination Healthcare for African Settings

# Evaluating Fine-Tuned and Base Language Models in Maternal and Vaccination Healthcare for African Settings

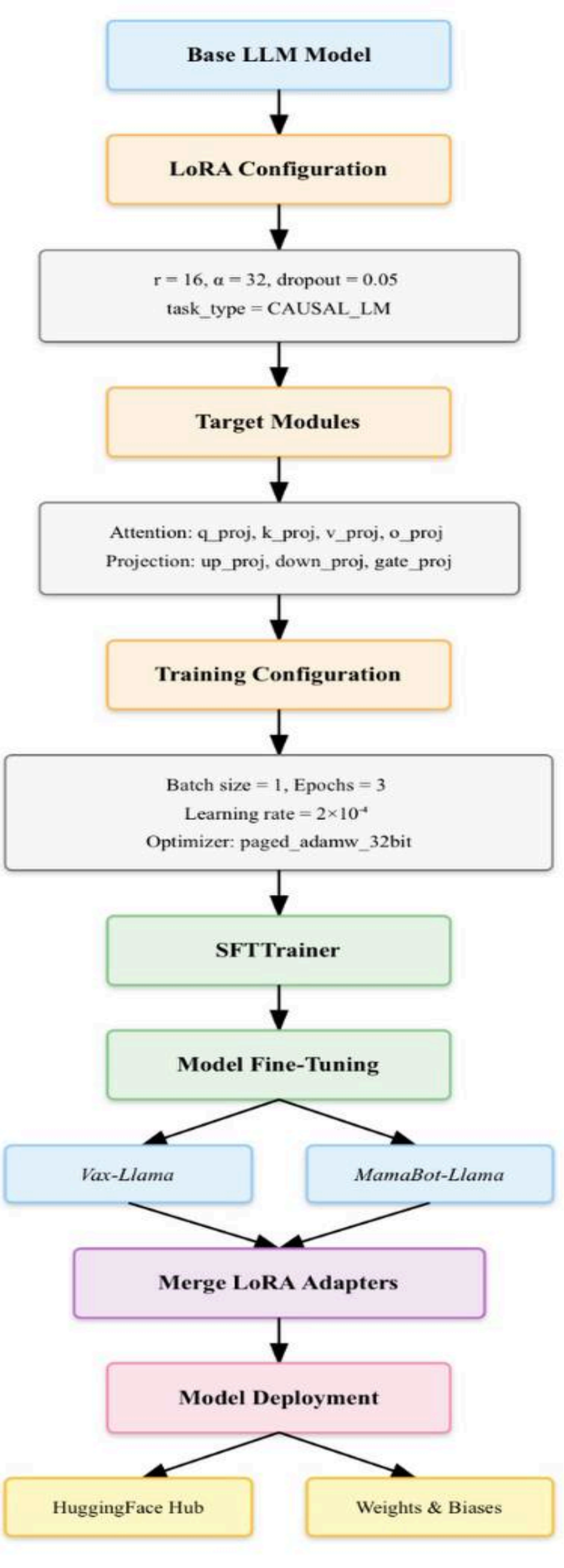

# Evaluating Fine-Tuned and Base Language Models in Maternal and Vaccination Healthcare for African Settings

**Figure 2:** *Fine-tuning architecture using Low-Rank Adaptation (LoRA). The base Llama-3.1-8B-Instruct model is augmented with trainable LoRA adapters (r=16, alpha=32) on attention and projection layers, producing domain-specific models (MamaBot-Llama, Vax-Llama).*

## F Evaluation Framework

### 1. Question Bank Development

A comprehensive question bank consisting of 200 questions (100 per domain) was created and stratified into five subdomains with 20 questions each. Maternal health subdomains included prenatal care, nutrition during pregnancy, pregnancy complications, postpartum care, and newborn care. Vaccination subdomains encompassed vaccine schedules, vaccine safety, vaccine efficacy, special vaccination cases, and disease prevention strategies.

Questions were generated using GPT-4o and Claude 3.5 Sonnet to ensure diversity and relevance, specifically reflecting real-world healthcare inquiries common in African healthcare settings.

### 2. Response Generation and Evaluation Protocol

Each question was presented to both the base model and the corresponding fine-tuned model (MamaBot-Llama for maternal health questions, Vax-Llama for vaccination questions). All responses were anonymized and presented to evaluators in a blinded format to minimize assessment bias.

The two Nigerian licensed physicians independently evaluated all responses using a standardized 5-point Likert scale (1 = *Strongly Disagree*, 5 = *Strongly Agree*) across five criteria:

1. Medical accuracy of information provided
2. Safety and appropriateness for patient care

3. Appropriateness for African healthcare contexts
4. Clarity and ease of understanding
5. Clinical trustworthiness

Evaluators were instructed to flag responses containing medical errors, safety concerns, or contextual inappropriateness, with optional qualitative comments. Evaluations were carried out relying on expert clinical judgment.

## G. Statistical Analysis

Mean Likert scale ratings were calculated for each evaluation criterion and model type. The Wilcoxon signed-rank test was used to assess statistical significance of differences between base and fine-tuned models, chosen for its appropriateness with paired, non-parametric data (Wilcoxon, 1945). Critical issue counts were tabulated and analyzed descriptively.

We assumed a non-normal distribution for our Likert scale data, consistent with established methodological research showing that ordinal data with limited response categories rarely meet normality assumptions (Norman, 2010; Sullivan & Artino, 2013; Jamieson, 2004). Accordingly, non-parametric statistical tests were used throughout.

Because ten separate comparisons were conducted (five criteria × two domains), no formal adjustment for multiple comparisons (e.g., Bonferroni) was applied. This decision follows the recommendation that such adjustments are often overly conservative for exploratory analyses with conceptually distinct outcomes (Rothman, 1990). However, p-values near the 0.05 threshold (e.g., $p = 0.012$ for African context relevance) should be interpreted with caution.

Sample size was determined using power analysis for a paired t-test as a conservative approximation for the Wilcoxon signed-rank test, which retains approximately 95% of the t-test's power for moderate sample sizes (Fagerland, 2012). A small-to-medium effect size (Cohen's $d = 0.3$) was targeted, consistent with clinically meaningful differences in healthcare evaluation studies (Cohen, 1988; Lakens, 2022). A priori power analysis indicated that 90 paired observations would provide 80% power to detect a small-to-medium effect ($d = 0.3$) at $\alpha = 0.05$. We increased our sample to 200 questions to ensure robustness against potential rater variability and to enable detection of smaller effects.

# Evaluating Fine-Tuned and Base Language Models in Maternal and Vaccination Healthcare for African Settings

All statistical analyses were conducted using Python with the following libraries: Pandas for data manipulation, NumPy for numerical operations, SciPy for statistical testing, and Matplotlib/Seaborn for data visualization including bar charts and radar plots. Inter-rater reliability was assessed using Cohen's κ to ensure evaluator consistency.

## H. Ethical Considerations

Informed consent for evaluators: The two physician evaluators (co-authors IA and CN) provided verbal informed consent before participating. The principal investigator explained the study purpose, the evaluation task, the use of a standardized 5-point Likert scale, the fact that aggregated ratings would be published in a peer-reviewed journal, and their right to withdraw at any time without consequence. No written consent form was used, given the minimal-risk, co-investigator nature of the activity. Both evaluators confirmed their understanding and agreement verbally. Their individual identities are known to the research team, but their ratings were anonymized during data aggregation and are presented without attribution to specific individuals.

Anonymization of chatbot logs: Conversation logs from MamaBot AI and Vax-AI were anonymized by: (a) dropping all columns containing direct user identifiers (user ID, phone number, name, location), (b) retaining only the cleaned question-answer pairs. After this process, no re-identification was possible.

Dual role disclosure: The physician evaluators are listed as co-authors because they contributed substantially to data acquisition (evaluation of model responses) and reviewed the final manuscript. Their role as evaluators does not compromise the integrity of the blinded assessment, as they were blinded to model identity (fine-tuned vs. base) during the rating process.

Data security: All data were stored on password-protected servers accessible only to the research team. No patient medical records or protected health information were used.

# Evaluating Fine-Tuned and Base Language Models in Maternal and Vaccination Healthcare for African Settings

# IV. Results

## A. Evaluation Overview

A total of 400 evaluations were completed across both domains, with 200 evaluations per domain covering 100 unique questions for each evaluator, per domain. Inter-rater reliability was excellent (Cohen's $\kappa = 0.82$).

## B. Maternal Health Domain Performance

### 1. Quantitative Performance Analysis

The fine-tuned MamaBot-Llama model demonstrated superior performance compared to the base Llama-3.1-8B-Instruct model across all five evaluation criteria (see Table 1). Statistical analysis revealed significant improvements.

**Table 1**

*Performance Comparison Between Base Model (Llama-3.1-8B-Instruct) and MamaBot-Llama in Maternal Health Domain*

| Criterion | **Base Model** Llama-3.1-8B-Instruct | MamaBot-Llama | Difference | *p* |
|---|---|---|---|---|
| Medical Accuracy | 4.17 | 4.38 | +0.21 (+5%) | .003 |

| Safety/Appropriateness | 4.12 | 4.29 | +0.17 (+4%) | .008 |
|---|---|---|---|---|
| African Context Relevance | 3.95 | 4.10 | +0.15 (+4%) | .012 |
| Clarity | 4.04 | 4.19 | +0.15 (+4%) | .011 |
| Clinical Trustworthiness | 4.09 | 4.38 | +0.29 (+7%) | < .001 |
| Overall Performance | 4.07 | 4.27 | +0.20 (+5%) | < .001 |

The most substantial improvement was observed in clinical trustworthiness ($p < .001$), followed by medical accuracy ($p = .003$). These findings suggest that domain-specific fine-tuning enhanced the model's reliability and clinical applicability in maternal health contexts.

# Evaluating Fine-Tuned and Base Language Models in Maternal and Vaccination Healthcare for African Settings

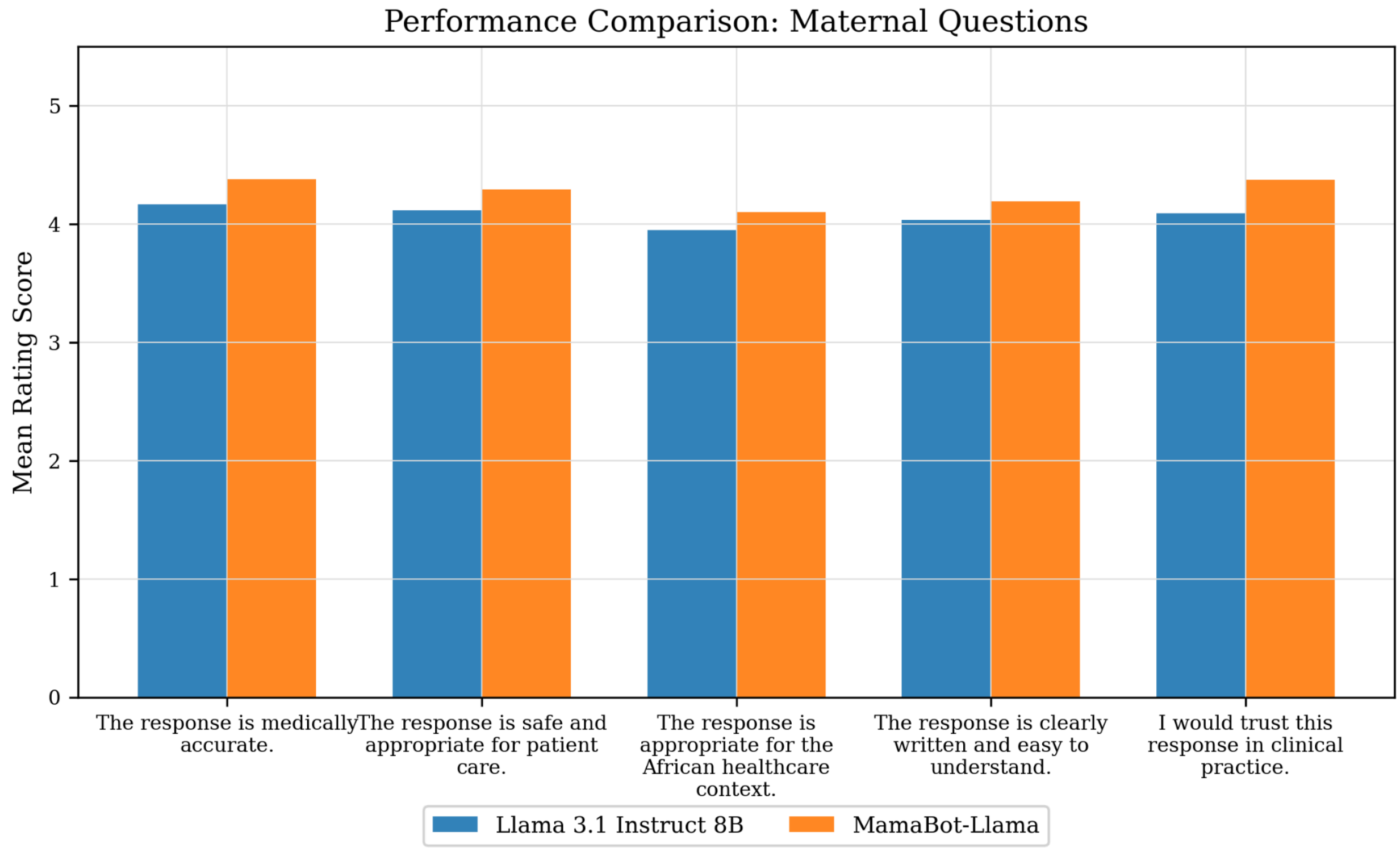

**Figure 3**

*Bar Plot Showing Performance Comparison Between Base Model (Llama 3.1 Instruct 8B) and MamaBot-Llama in Maternal Health Domain*

### 2. Critical Issue Analysis

Analysis of critical issues revealed a marked reduction in problematic responses for the fine-tuned model (see Table 2 below). The MamaBot-Llama model exhibited 50% fewer total critical issues compared to the base model.

# Evaluating Fine-Tuned and Base Language Models in Maternal and Vaccination Healthcare for African Settings

**Table 2**

*Critical Issues by Model Type in Maternal Health Domain*

| Issue Category | Base Model Llama-3.1-8B-Instruct | MamaBot-Llama | Reduction |
|---|---|---|---|
| Medical Errors | 4 | 3 | 25% |
| Safety Concerns | 8 | 3 | 62.5% |
| Inappropriate Context | 2 | 1 | 50% |
| Total | 14 | 7 | 50% |

The most notable improvement was in safety concerns, with a 62.5% reduction in potentially harmful recommendations. One evaluator noted that "the base model recommended ultrasound frequency impractical for rural clinics," highlighting the improved contextual appropriateness of the fine-tuned model.

# Evaluating Fine-Tuned and Base Language Models in Maternal and Vaccination Healthcare for African Settings

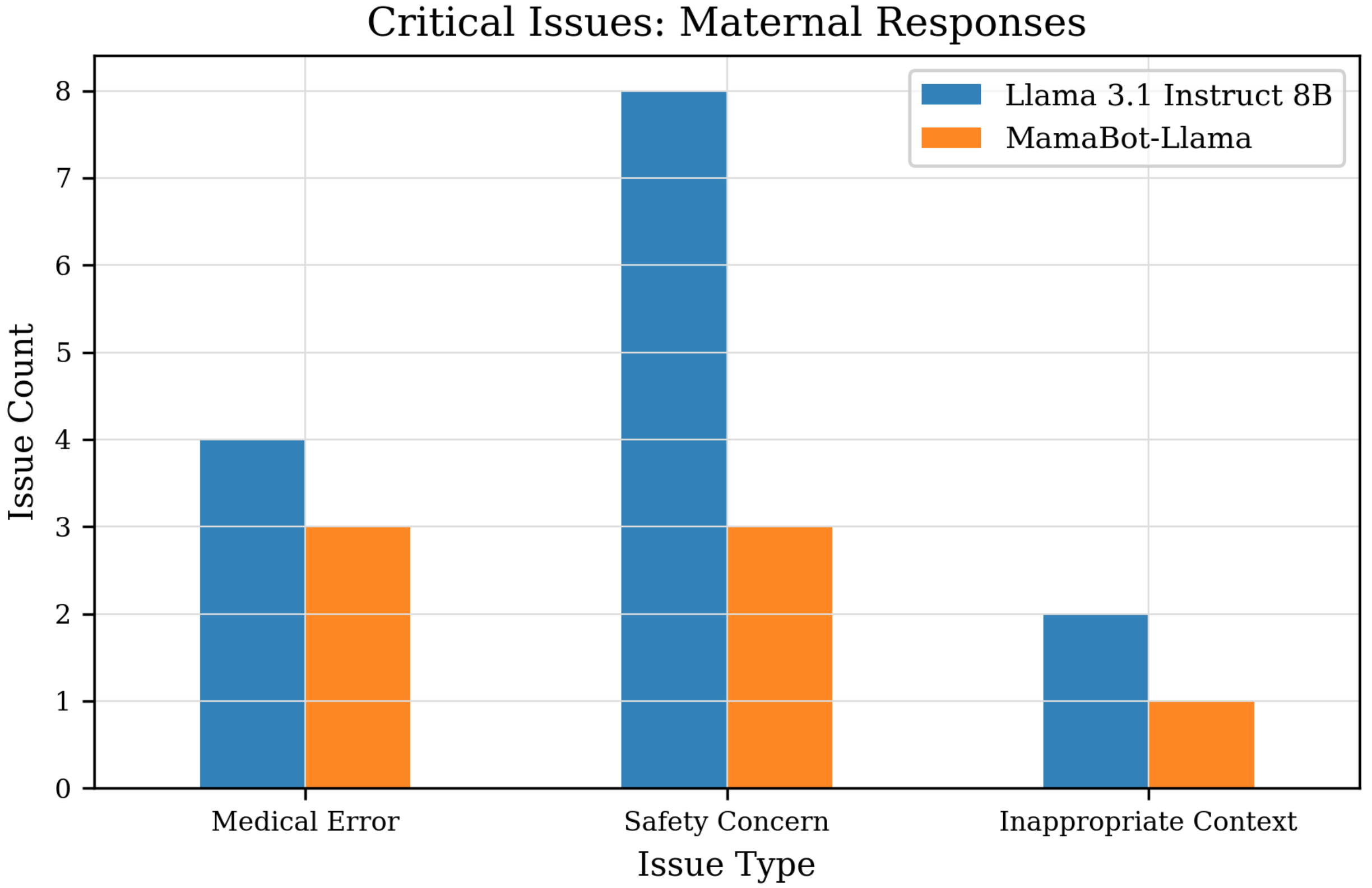

**Figure 4**

*Bar Plot Showing Critical Issues by Model Type in Maternal Health Domain*

# Evaluating Fine-Tuned and Base Language Models in Maternal and Vaccination Healthcare for African Settings

## 3. Clinician Preference Analysis

Evaluator preferences strongly favored the MamaBot-Llama model, with clinicians preferring it in 78% of maternal health cases. The mean preference score for MamaBot-Llama was significantly higher ($M$ = 4.47) compared to the base model ($M$ = 4.11), indicating substantial clinical preference for the domain-specific model.

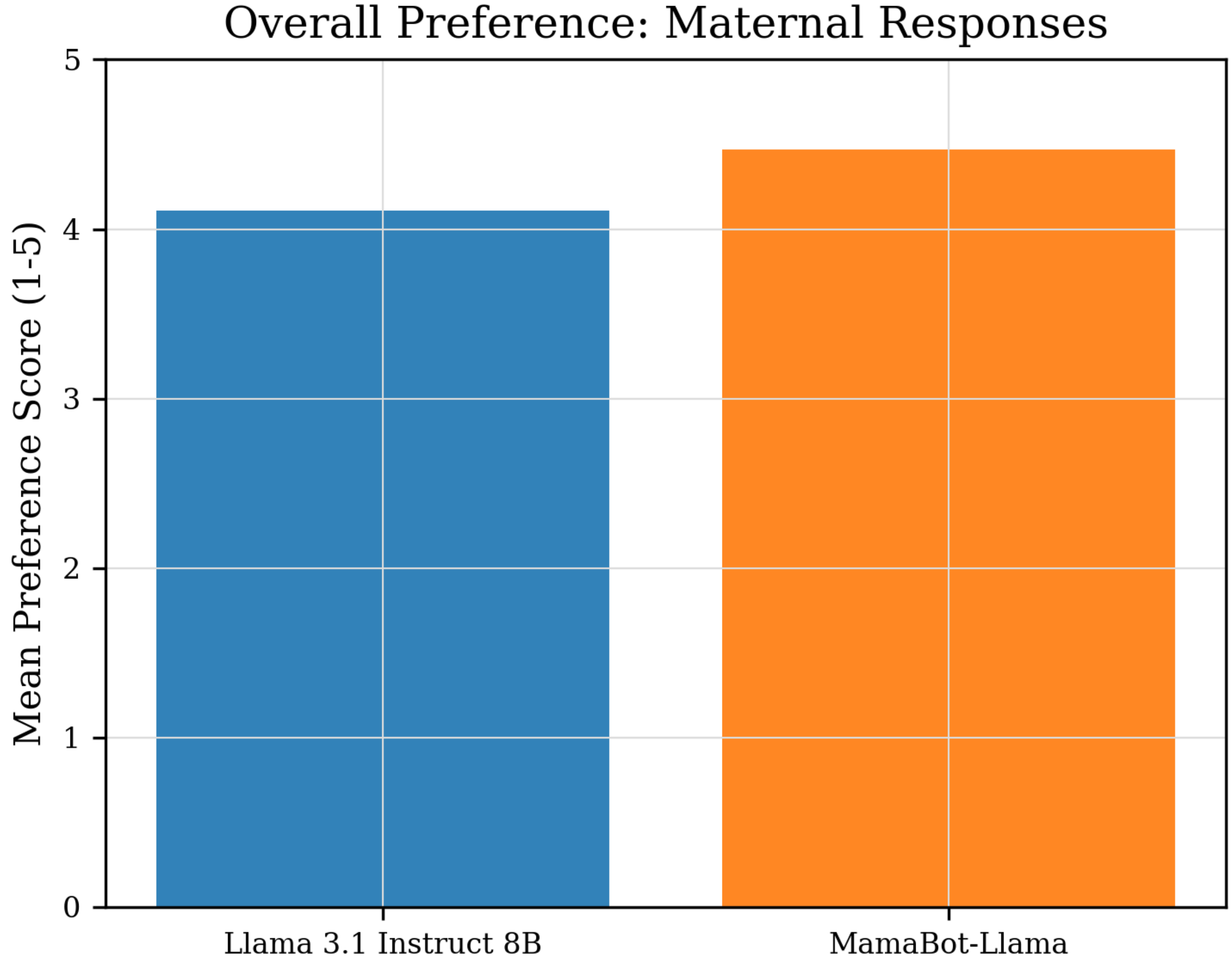

**Figure 5**

*Bar Plot Showing Clinician Preference by Model Type in Maternal Health Domain*

## C. Vaccination Domain Performance

# Evaluating Fine-Tuned and Base Language Models in Maternal and Vaccination Healthcare for African Settings

## 1. Quantitative Performance Analysis

Contrary to the maternal health domain, the base Llama-3.1-8B-Instruct model outperformed the fine-tuned HelpMum Vax-Llama model across all evaluation criteria (see Table 3). This finding suggests that fine-tuning may have introduced domain-specific limitations in vaccination-related responses.

**Table 3**

*Performance Comparison Between Base Model (Llama-3.1-8B-Instruct) and Vax-Llama in Vaccination Domain*

| **Criterion** | **Base Model** Llama-3.1-8B-Instruct | **Vax-Llama** | **Difference** | ***p*** |
|---|---|---|---|---|
| Medical Accuracy | 4.57 | 4.34 | -0.23 (-5%) | .004 |
| Safety/Appropriateness | 4.29 | 3.98 | -0.31 (-7%) | < .001 |
| African Context Relevance | 3.99 | 3.79 | -0.20 (-5%) | .009 |
| Clarity | 4.10 | 3.96 | -0.14 (-3%) | .021 |
| Clinical Trustworthiness | 4.28 | 4.07 | -0.21 (-5%) | .002 |

| Overall Performance | 4.25 | 4.03 | -0.22 (-5%) | < .001 |
|---|---|---|---|---|

The largest performance gap was observed in safety and appropriateness ($p < .001$), raising concerns about the clinical safety of the fine-tuned vaccination model.

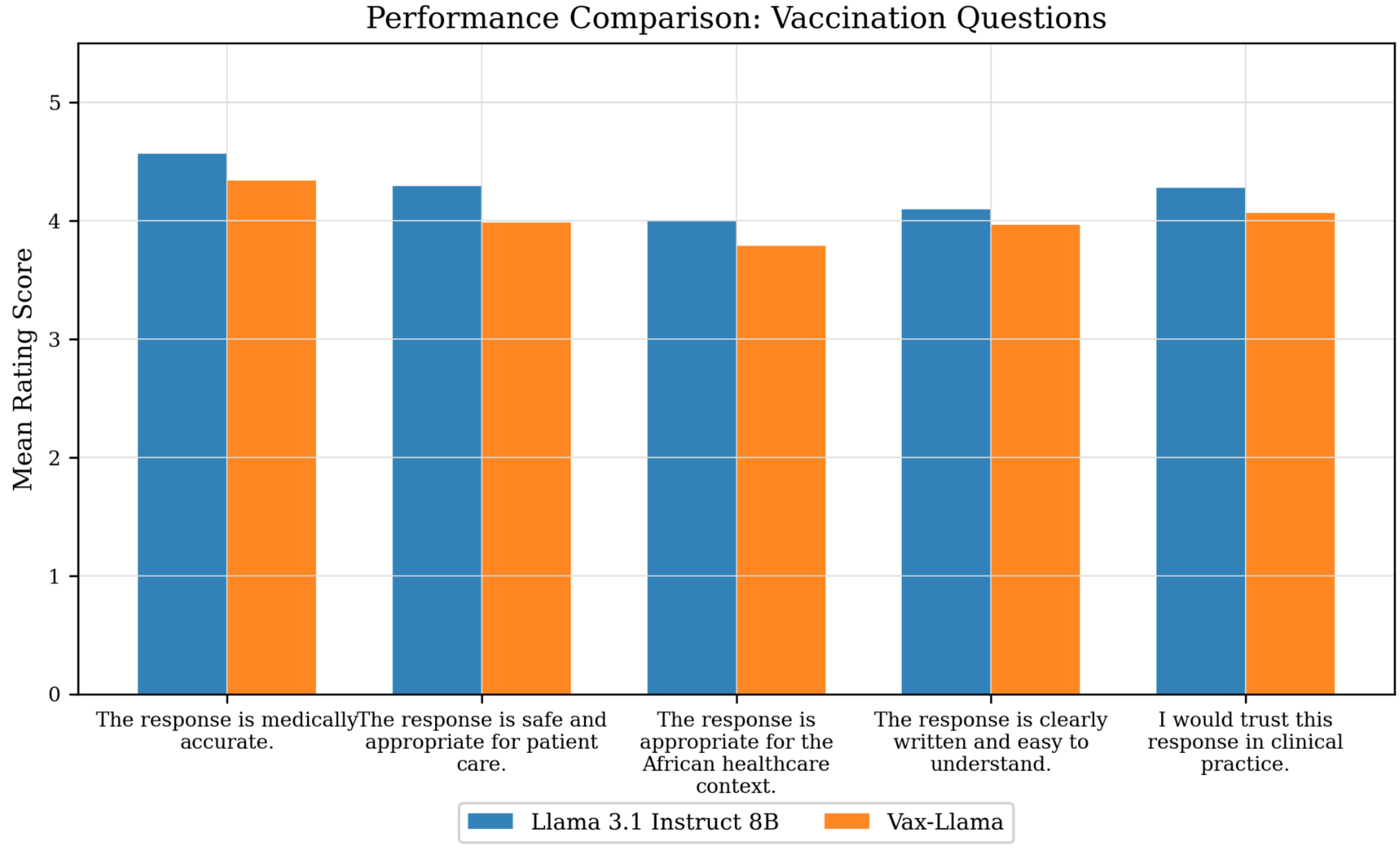

**Figure 6**

*Bar Plot Showing Performance Comparison Between Base Model (Llama 3.1 Instruct 8B) and Vax-Llama in Vaccination Domain*

## 2. Critical Issue Analysis

The vaccination domain analysis revealed a concerning increase in critical issues for the fine-tuned model (see Table 4). The Vax-Llama model exhibited nearly three times more critical issues than the base model (38 vs. 13 instances).

# Evaluating Fine-Tuned and Base Language Models in Maternal and Vaccination Healthcare for African Settings

**Table 4**

*Critical Issues by Model Type in Vaccination Domain*

| Issue Category | Base Model<br>Llama-3.1-8B-Instruct | Vax-Llama | Increase |
|---|---|---|---|
| Medical Errors | 4 | 7 | 75% |
| Safety Concerns | 3 | 15 | 400% |
| Inappropriate Context | 6 | 16 | 167% |
| Total | 13 | 38 | 192% |

# Evaluating Fine-Tuned and Base Language Models in Maternal and Vaccination Healthcare for African Settings

The most alarming finding was the 400% increase in safety concerns, with 68% of these issues involving inappropriate cold chain recommendations. One evaluator specifically noted that the "Vax-Llama model underestimated reconstitution time constraints in power-outage scenarios," highlighting potential real-world safety implications.

# Evaluating Fine-Tuned and Base Language Models in Maternal and Vaccination Healthcare for African Settings

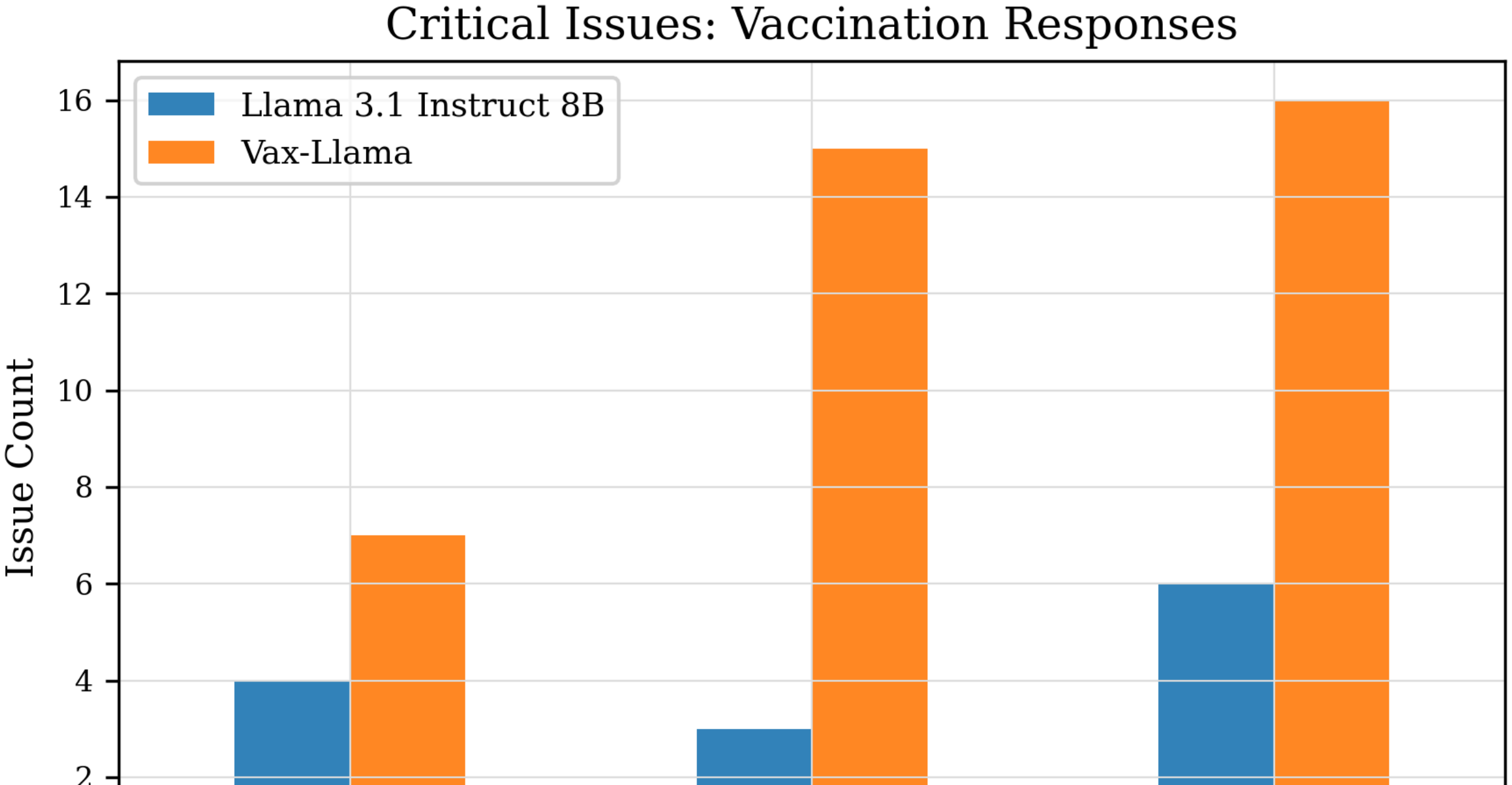

**Figure 7**

*Bar Plot Showing Critical Issues by Model Type in Vaccination Domain*

## 3. Clinician Preference Analysis

Despite the quantitative performance differences, clinician preferences did not differ significantly between models (Vax-Llama M = 4.16 vs. base model M = 4.14; Wilcoxon signed-rank test, $p = .32$). The near-equivalent scores indicate that, while the fine-tuned model scored lower on individual criteria, it was not globally disfavored. However, given the non-significant p-value, no conclusion about qualitative appeal is warranted.

# Evaluating Fine-Tuned and Base Language Models in Maternal and Vaccination Healthcare for African Settings

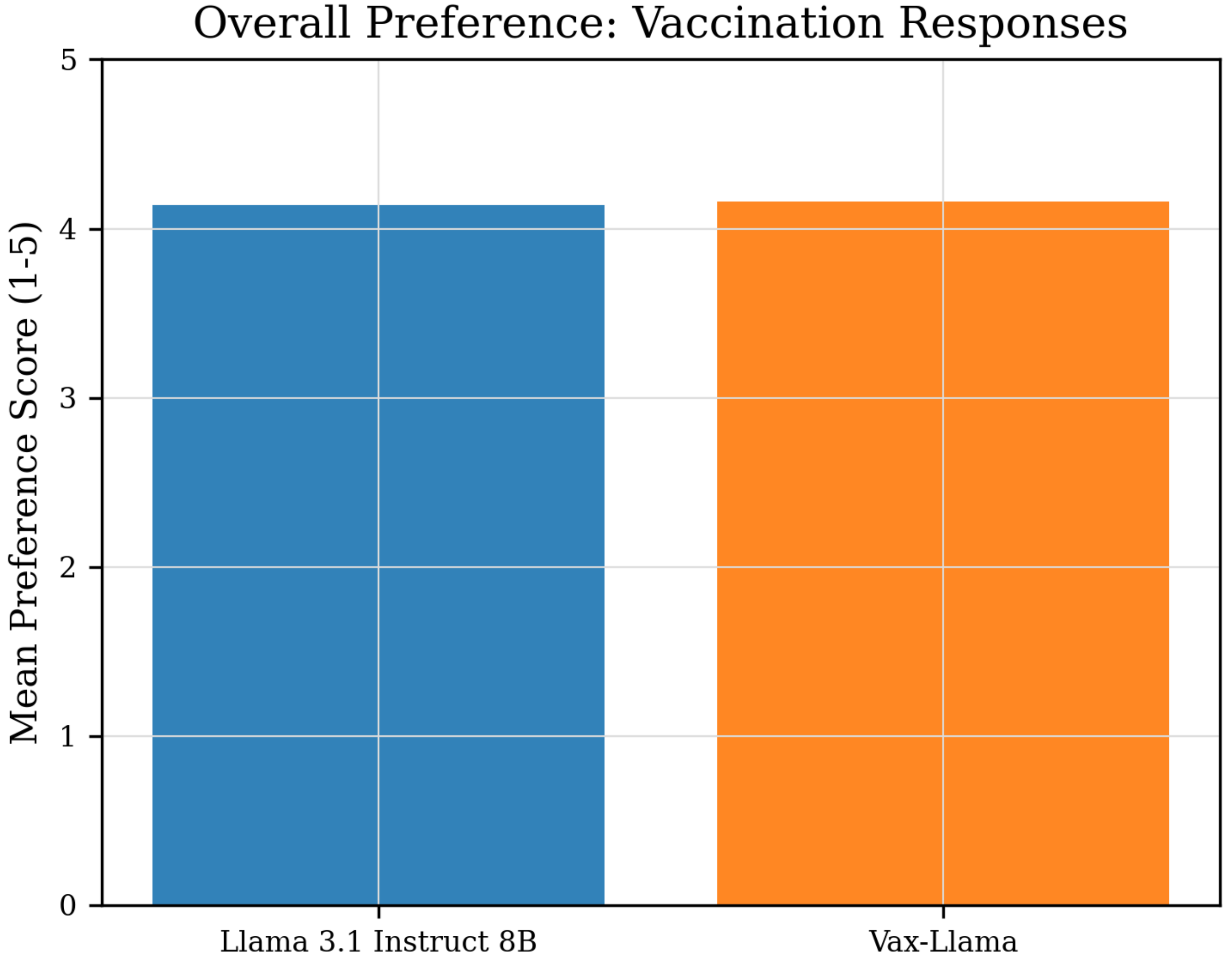

**FIgure 8**

*Bar Plot Showing Clinician Preference by Model Type in Vaccination Domain*

## D. Cross-Domain Performance Comparison

Comparative analysis across domains revealed divergent outcomes for fine-tuning effectiveness. In the maternal health domain, fine-tuning resulted in a 4.9% improvement in overall performance (base model: 4.07, MamaBot-Llama: 4.27). Conversely, in the vaccination domain, fine-tuning led to a 5.2% decrease in overall performance (base model: 4.25, Vax-Llama: 4.03).

These contrasting results may be attributed to several factors. A notable factor is the quality of the dataset used, approximately 89% of the vaccination dataset was created using the LLM-augmented web-scrapping method, with no expert or clinician created data included, unlike the maternal health data which has approximately 32% of clinician-curated content, enhancing overall performance.

Other factors include dataset quantity differences (maternal health: 36,000+ training examples vs. vaccination: 9,000+ training examples) and domain complexity variations.

# V. Discussion

The integration of large language models (LLMs) into healthcare offers transformative potential, particularly in low-resource settings where access to accurate and contextually relevant health information is limited. This study evaluated the performance of two domain-specific fine-tuned LLMs, MamaBot-Llama for maternal health and Vax-Llama for vaccination, against the base model, Meta's Llama-3.1-8B-Instruct, in delivering high-quality responses tailored to Nigerian contexts. The findings reveal a striking contrast: fine-tuning significantly enhanced performance in the maternal health domain but led to a decline in the vaccination domain. These results highlight both the promise and the challenges of adapting LLMs for healthcare applications in underserved regions.

## A. Summary of Key Findings

In the maternal health domain, the MamaBot-Llama model outperformed the base model across all evaluation criteria: medical accuracy, safety and appropriateness, African context relevance, clarity, and clinical trustworthiness. The overall performance improved by 4.9%, with notable gains in clinical trustworthiness (+7%) and medical accuracy (+5%). Critical issues, including safety concerns, decreased by 50%, and clinicians preferred the fine-tuned model in 78% of cases. These improvements suggest that fine-tuning can enhance the reliability and relevance of AI-generated maternal health advice.

Conversely, in the vaccination domain, the Vax-Llama model underperformed the base model, with an overall performance decline of 5.2%. The most significant reductions were in safety and appropriateness (-7%) and medical accuracy (-5%), accompanied by a 192% increase in critical issues, including a 400% rise in safety concerns. Despite this, clinicians showed a marginal preference for the fine-tuned model, hinting at qualitative strengths not captured by quantitative metrics. These divergent outcomes underscore the variable impact of fine-tuning across healthcare domains.

## B. Interpretation of Results

The success of fine-tuning in the maternal health domain likely stems from the quality and size of the training dataset. With over 36,000 question-answer pairs, including 32% clinician-curated content, the maternal health dataset provided a robust foundation for enhancing the model's performance. This aligns with prior research indicating that high-quality, domain-specific data can improve LLM accuracy and safety (Singhal et al., 2023). The reduction in safety concerns further suggests that fine-tuning improved contextual appropriateness, a critical factor in low-resource settings like Nigeria (EQUAL Research Consortium, 2023).

In contrast, the vaccination domain's poorer performance may be attributed to a smaller dataset (over 9,000 pairs) with a higher proportion of LLM-Augmented content (89%) and less clinician oversight. This disparity likely introduced noise or biases during fine-tuning, leading to errors such as inappropriate cold chain recommendations. Vaccination is a complex domain, requiring precise guidance on logistics (e.g., reconstitution times) and local epidemiology, which may have been inadequately represented in the training data. The increase in safety concerns echoes warnings about LLMs amplifying errors when fine-tuned on suboptimal datasets (Alkaissi & McFarlane, 2023), emphasizing the need for rigorous data curation.

The paradoxical clinician preference for Vax-Llama, despite its lower ratings, suggests that factors like language style or perceived empathy may influence subjective judgments. This highlights a tension between quantitative performance and qualitative appeal, a phenomenon warranting further exploration in AI healthcare applications (Presiado et al., 2024).

## C. Implications for AI in Healthcare

These findings have significant implications for deploying LLMs in low-resource healthcare settings. In maternal health, the success of MamaBot-Llama suggests that fine-tuned models can serve as valuable tools for delivering accurate and culturally relevant advice, potentially improving health literacy and outcomes in regions with high maternal mortality rates (Bello et al., 2022). This is particularly promising for Nigeria, where scalable solutions could address gaps in antenatal care access.

However, the vaccination domain results caution against overgeneralizing the benefits of fine-tuning. The increase in critical issues, especially safety concerns, underscores the risk of deploying inadequately validated models in clinical contexts. Inaccurate vaccination advice could exacerbate hesitancy or lead to harmful practices, undermining public health efforts (Adeyanju et al., 2022). This highlights the need for domain-specific evaluation frameworks that prioritize safety and accuracy over other metrics.

The study also emphasizes the importance of cultural and contextual adaptation. Generic LLMs may provide advice misaligned with local realities, such as recommending unavailable diagnostics. Fine-tuning with region-specific data, as seen in maternal health, can mitigate this, enhancing trust and usability among users (Presiado et al., 2024). However, achieving this requires investment in high-quality, localized datasets, a challenge in resource-constrained environments.

## D. Limitations of the Study

Some limitations temper these findings. First, the evaluation relied on a modest question set (100 per domain) assessed by only two Nigerian physicians. Although inter-rater reliability was high (Cohen's $\kappa = 0.82$), a larger and more diverse panel, including midwives or community health workers, could strengthen the results' generalizability. Second, the datasets differed significantly in size and quality, confounding direct comparisons between domains. The maternal health dataset's richness likely drove its success, while the vaccination dataset's limitations may have hindered performance.

Additionally, a substantial portion of the training data (89% of the vaccination dataset, and an unknown fraction of the maternal health dataset from web scraping) was generated by using GPT-4o to convert raw scraped text into question-answer pairs. This introduces a risk of model-specific bias and circularity: the fine-tuned models were optimized on data derived from another LLM. The poor performance of Vax-Llama may partly reflect this; the maternal health dataset's inclusion of 32% clinician-curated content likely mitigated the issue. Future work should prioritize human-curated or clinically verified datasets over LLM-generated synthetic pairs.

Also, the study lacked patient perspectives and real-world testing. Understanding how Nigerian mothers interpret and act on AI advice is critical for assessing its practical utility.

## E. Ethical Considerations and Broader Context

Deploying LLMs in healthcare raises ethical challenges, particularly in low-resource settings. Data privacy is paramount, as health queries involve sensitive information. Compliance with regulations and robust anonymization are essential to protect users. Additionally, AI risks exacerbating inequalities if access is limited to digitally literate populations, necessitating equitable deployment strategies.

Excessive reliance on AI advice is another concern. While LLMs can supplement knowledge, they should not replace professional care. In Nigeria, where health literacy is low, users may misinterpret confidently delivered errors, necessitating clear disclaimers about the AI's limitations (Alkaissi & McFarlane, 2023). Balancing innovation with safety remains a critical ethical imperative.

## F. Future Research Directions

Future studies should address these limitations and build on these findings. First, refining fine-tuning techniques, such as reinforcement learning with human feedback, could optimize performance across domains. Second, expanding dataset size and diversity, particularly for vaccination, with more clinician-curated content, may mitigate errors and biases. Third, incorporating patient feedback and conducting field trials in Nigerian communities would clarify the models' real-world impact on health behaviors and outcomes.

Additionally, evaluating multilingual capabilities, including Nigerian Pidgin English and local languages, could enhance accessibility. Developing comprehensive evaluation frameworks that integrate quantitative metrics (e.g., accuracy) with qualitative factors (e.g., empathy) would provide a fuller picture of model suitability. Finally, longitudinal studies assessing the long-term effects of AI-driven health tools on maternal and child health metrics could guide scalable implementation.

# VI. Conclusion

This study provides critical insights into the effectiveness of domain-specific fine-tuning for large language models in healthcare applications within low-resource settings. The findings reveal that fine-tuning outcomes are highly domain-dependent, with significant implications for the deployment of AI-driven health information systems in underserved communities like Nigeria.

The success of the MamaBot-Llama model in maternal health demonstrates the transformative potential of carefully curated, domain-specific fine-tuning. With a 4.9% overall performance improvement, 50% reduction in critical issues, and strong clinician preference (78%), this model offers a promising pathway for addressing Nigeria's maternal mortality crisis through accessible, culturally relevant health information. The substantial gains in clinical trustworthiness and medical accuracy suggest that well-executed fine-tuning can enhance both the reliability and contextual appropriateness of AI-generated health advice.

Conversely, the Vax-Llama model's underperformance in the vaccination domain serves as a cautionary tale about the risks of inadequate fine-tuning. The 5.2% performance decline and alarming 192% increase in critical issues, particularly safety concerns, underscore the potential dangers of deploying poorly validated models in clinical contexts. These findings highlight that dataset quality, size, and clinician oversight are paramount to successful fine-tuning outcomes.

The study's implications extend beyond technical considerations to fundamental questions about AI governance in healthcare. The contrasting results emphasize the necessity of rigorous, domain-specific evaluation frameworks before clinical deployment, particularly in settings where healthcare access is limited and the consequences of misinformation can be severe. Future research should prioritize comprehensive dataset curation, multi-stakeholder evaluation approaches, and real-world testing to ensure that AI-driven health tools genuinely serve the needs of vulnerable populations while maintaining the highest standards of safety and accuracy.

# List of Abbreviations

- LLM: Large Language Model
- LoRA: Low-Rank Adaptation
- WHO: World Health Organization
- UNICEF: United Nations Children's Fund
- CDC: Centers for Disease Control and Prevention
- API: Application Programming Interface

# Evaluating Fine-Tuned and Base Language Models in Maternal and Vaccination Healthcare for African Settings

- GPT: Generative Pre-trained Transformer
- AI: Artificial Intelligence
- HTML: Hypertext Markup Language
- CSS: Cascading Style Sheets
- JSON: JavaScript Object Notation
- URL: Uniform Resource Locator

## Declarations

### Ethics Approval and Consent to Participate

The work was conducted as a minimal-risk technical evaluation of AI models using: (i) two physician co-authors who assessed model outputs as part of their role on the research team, and (ii) anonymized, previously collected chatbot interaction logs.

Both physician evaluators provided verbal informed consent prior to participation, as described in Section III.H. No written consent form was used, consistent with the minimal-risk, co-investigator nature of the activity.

The chatbot conversation logs (from HelpMum's MamaBot AI and Vax-AI) were anonymized before analysis by removing all user-identifiable fields (e.g., phone numbers, user IDs, names). Use of these logs for research is permitted under HelpMum's terms of service, which state that anonymized interaction data may be used for service improvement and research. No identifiable patient data or medical records were used.

### Consent for Publication

Not applicable. This manuscript does not contain data from any individual person.

# Evaluating Fine-Tuned and Base Language Models in Maternal and Vaccination Healthcare for African Settings

## Availability of Data and Materials

The evaluation question banks and model response datasets generated during the current study are available from the corresponding author on reasonable request. The fine-tuned models (MamaBot-Llama and Vax-Llama) are proprietary to HelpMum and available at https://helpmum.org/mamabotllama and https://helpmum.org/vax_llama respectively. The base model (Meta-Llama-3.1-8B-Instruct) is publicly available from Meta AI at https://ai.meta.com/llama/. Training datasets were derived from publicly available sources (WHO, UNICEF, CDC Nigeria) and anonymized chatbot conversation logs, with web scraping code and preprocessing scripts available upon reasonable request.

## Competing Interests

The authors declare that they have no competing interests. AJIBADE, A., ODUNSI, O., ANIMASAUN, I., NWAKANMA-AKANNO, C., OGUNTUASE, O., and ADERENI, A. are affiliated with HelpMum, the organization that developed the fine-tuned models evaluated in this study. However, the evaluation was conducted by independent Nigerian licensed physicians who were blinded to model identities, and statistical analyses were performed objectively without bias toward either model type.

## Funding

This research received no specific grant from any funding agency in the public, commercial, or not-for-profit sectors.

## Authors' Contributions

- Abdulquddus Ajibade (AA): Conceptualization, model fine-tuning and training, response generation, data preprocessing, statistical analysis, interpretation, manuscript drafting.
- Oluwaseun Odunsi (OO): Data preprocessing, manuscript editing.
- Iyinoluwa Animasaun (IA): Physician evaluator (blinded), performed all ratings for maternal health and vaccination responses.
- Chioma Nwakanma-Akanno (CN): Physician evaluator (blinded), performed all ratings for maternal health and vaccination responses.

- Oluwasegun Oguntuase (OOg): Manuscript writing and editing (note: different from Odunsi).
- Abiodun Adereni (AA): Supervision, study design guidance, manuscript editing.

All authors read and approved the final manuscript. The two physician evaluators (IA, CN) are co-authors; their dual role is disclosed. The statistical analysis was conducted by Ajibade, who was not involved in rating.

## Acknowledgements

The authors wish to thank CN and IA for their invaluable time and clinical expertise as the independent medical evaluators who assessed all model responses in this study. We acknowledge the HelpMum team for providing access to anonymized chatbot conversation logs from MamaBot AI and Vax-AI, and for technical infrastructure support during model training and evaluation. We are grateful to Meta AI for making the Llama 3.1 base model publicly available for research purposes. We also thank the World Health Organization (WHO), UNICEF, and Nigeria Primary Health Care Development Agency for maintaining publicly accessible health resources that contributed to our training datasets.

## Authors' Information

**Abdulquddus Ajibade** (MBBS) is Lead AI Engineer at HelpMum, medical doctor, Nigeria, and co-founder and CTO of OpenHealth. He specializes in AI/ML engineering and AI-driven health tools, with research interests in healthcare artificial intelligence applications.

**Oluwaseun Odunsi** is Executive Director at HelpMum and an Information Resource Manager with expertise in leveraging technology to drive organizational growth and effective information management in healthcare settings.

**Iyinoluwa Animasaun (MBBS)** is a Nigerian licensed physician working as medical personnel at HelpMum, with clinical interests in maternal and child health and experience in evaluating AI-driven healthcare interventions.

# Evaluating Fine-Tuned and Base Language Models in Maternal and Vaccination Healthcare for African Settings

**Chioma Nwakanma-Akanno (MBBS)** is Medical Director at HelpMum and a reproductive healthcare physician. She is the founder of Medically Speaking Services and Executive Director of the SMILE WITH Me Foundation, with extensive experience in maternal health advocacy and clinical practice.

**Oluwasegun Oguntuase** is Frontend Developer at HelpMum and a software engineer with previous experience at Aladdin Digital Bank and Perxels, specializing in healthcare technology interfaces.

**Abiodun Adereni (DVM)** is Founder and CEO of HelpMum and founder of Dobic Health. He is a vet-medical professional and social entrepreneur who leverages technology including AI-powered vaccination trackers and clean birth kits to improve healthcare delivery in underserved communities.

# Evaluating Fine-Tuned and Base Language Models in Maternal and Vaccination Healthcare for African Settings

# Evaluating Fine-Tuned and Base Language Models in Maternal and Vaccination Healthcare for African Settings

# APPENDIX

## EVALUATION FORM

**Evaluator ID: _______**
**Category: [Maternal Health / Vaccination]**
**Question : _______**

1. **Model Responses:**

Please rate your agreement with the following statements for each model's response using:
1 = Strongly Disagree, 2 = Disagree, 3 = Neutral, 4 = Agree, 5 = Strongly Agree

| **Criteria** | **Model A** | **Model B** |
|---|---|---|
| **The response is medically accurate.** | | |

# Evaluating Fine-Tuned and Base Language Models in Maternal and Vaccination Healthcare for African Settings

| **The response is safe and appropriate for patient care.** | | |
|---|---|---|
| **The response is appropriate for the African healthcare context.** | | |
| **The response is clearly written and easy to understand.** | | |
| **I would trust this response in clinical practice.** | | |

2. **Critical Issues (if any):**

Check (✓) boxes if a response contains medical errors, safety concerns, or contextual inappropriateness.

| **Issue Type** | **Model A** | **Model B** |
|---|---|---|
| **Medical Error** | | |
| **Safety Concern** | | |
| **Inappropriate Context** | | |

3. **Brief Comments (Optional):**

Add optional brief notes to explain your ratings (e.g., strengths, flaws, cultural mismatches).

Model A: ________________
Model B: ________________

4. **Preferred Response**

Select the single best model based on overall quality and trustworthiness. **[ ]**